\documentclass{article}
\usepackage{amsmath}
\usepackage[Symbol]{upgreek}
\usepackage{microtype}
\usepackage{graphicx}
\usepackage{amsfonts}
\usepackage{amsmath}
\usepackage{subfigure}
\usepackage{standalone}
\usepackage{booktabs} 
\usepackage{enumitem}
\usepackage{mathtools}
\DeclarePairedDelimiter\ceil{\lceil}{\rceil}

\usepackage{hyperref}

\usepackage[accepted]{icml2021}

\icmltitlerunning{Cyclical Kernel Adaptive Metropolis}

\begin{document}

\twocolumn[
\icmltitle{Cyclical Kernel Adaptive Metropolis}



\icmlsetsymbol{equal}{*}

\begin{icmlauthorlist}
\icmlauthor{Jianan Canal Li}{equal,c,bee}
\icmlauthor{Yimeng Zeng}{equal,c}
\icmlauthor{Wentao Guo}{equal,c}
\end{icmlauthorlist}

\icmlaffiliation{c}{Department of Computer Science,}
\icmlaffiliation{bee}{Biological and Environmental Engineering, Cornell University}

\icmlcorrespondingauthor{Jianan Canal Li}{jl3789@cornell.edu}
\icmlcorrespondingauthor{Yimeng Zeng}{yz443@cornell.edu}
\icmlcorrespondingauthor{Wentao Guo}{wg247@cornell.edu}

\icmlkeywords{Machine Learning, ICML}

\vskip 0.3in
]



\printAffiliationsAndNotice{\icmlEqualContribution}

\begin{abstract}
 We propose cKAM, cyclical Kernel Adaptive Metropolis, which incorporates a cyclical stepsize scheme to allow control for exploration and sampling. We show that on a crafted bimodal distribution, existing Adaptive Metropolis type algorithms would fail to converge to the true posterior distribution. We point out that this is because adaptive samplers estimates the local/global covariance structure using past history of the chain, which will lead to adaptive algorithms be trapped in a local mode.  We demonstrate that cKAM encourages exploration of the posterior distribution and allows the sampler to escape from a local mode, while maintaining the high performance of adaptive methods.
\end{abstract}

\section{Introduction}

Bayesian inference is know for its ability for probabilistic modeling for data. Markov Chain Monte Carlo (MCMC) is a class of sampling methods where we obtain a sequence of samples from a probability distribution. Given a dataset $\mathcal{D} = \left\{ x_i \right\}_{i=1}^N$ and a model parametrized by $\theta$, MCMC samplers construct a Markov chain with the desired distribution as its stationary distribution, it is guaranteed to converge to the true distribution
\[
    \uppi(\theta) \propto \exp{\Big( \log{p(\mathcal{D}|\theta)} + \log{p(\theta)} \Big)},
\]

The choice of proposal $q(\cdot | \theta_t)$ for constructing the Markov chain is crucial for the performance of an MCMC sampler. To increase the efficiency of a sampler, methods for adapting the covariance structure of posterior base on the running history of the chain such as Adaptive-Metropolis (AM) \cite{Haario}, global adaptive scaling and other more sophisticated schemes \cite{Andrieu2008ATO} have been introduced, and the mixing and ergodicity of this type of algorithms has been well studied \cite{Roberts2007CouplingAE} \cite{AdaMCMC_TM}. However, these samplers are only useful for distributions that show high anisotropy. To allow adaptation in strongly non-linear target distributions where the directions of variation depends the current position $\theta_t$, Kernel Adaptive Metropolis-Hastings (KAM) is proposed \cite{sejdinovic2014kernel}. KAM adapts to the local covariance structure, since the proposed next state $\theta'$ is the pre-image of a sample in the Reproducing Kernel Hilbert Space (RKHS) with a Gaussian Measure. 

\par In this work, we propose a variant of KAM, the cyclic Kernel Adaptive Metropolis (cKAM), that shows local adaptation, while using cyclic stepsize scheme to control \textbf{Exploration} and \textbf{Sampling}. Compare with KAM, cKAM only recomputes kernel gradient matrix $M_{\mathbf{z},\theta}$ during the  \textbf{Exploration}, thus greatly reduces the cost of making a proposal. \par
Through experimenting with a synthetic two-dimensional bimodal target distribution,  we demonstrate that cKAM's stepsize scheme allows us to jump between different modes and converges to the correct distribution, whereas previous adaptive samplers failed in finding all the modes. We also showed that cKAM is able to retain similar performance as KAM in following non-linear targets: 1) 2d Gaussian Mixtures target distribution and 2) 32-dimensional Gaussian Mixtures target distribution. 


\section{Kernel Adaptive Metropolis-Hastings}
We restates algorithm formulation for KAM in Algorithm \ref{alg:KAM}, clearly identify adaptation for both stepsize and covariance. KAM adapts to the local covariance structure, by proposing the kernel form of next position, $f=k(\cdot, \theta')$, in a reproducing kernel Hilbert space (RKHS) $\mathcal{H}_k$ with a Gaussian measure. The ``density" form of this measure can be denoted as $\mathcal{N}(f; k(\cdot,\theta_t), \nu^2 C_\mathbf{z})$ where $C_\mathbf{z}$ is the empirical covariance operator on a random subsample of chain history $\mathbf{z}$. The pre-image $\theta'$ of $f$ is found by taking a gradient descent step w.r.t. $\theta_t$, and this move can be analytically integrated out, as in line 9, Algorithm \ref{alg:KAM}. 

We note that the KAM proposal is different from 1) Random-Walk (RW) proposal where $q(\cdot | \theta_t) = \mathcal{N}({\cdot | \theta_t, \Sigma})$ \cite{brooks2011handbook} and 2) gradient-based proposals using Hamiltonian dynamics \cite{Neal2011MCMCUH} or Langevin dynamics\cite{Roberts2002LangevinDA}. Although KAM proposal doesn't use gradient information of the posterior as it maps past chain history into RKHS space, it still captures some level of neighboring posterior geometry; therefore KAM's proposal lies in between RW and gradient-based proposals. 

There are two computational costs associated with making a proposal in RKHS space: 1) computing the kernel gradient matrices for both $M_{\mathbf{z},\theta'}^T$ and $M_{\mathbf{z},\theta_t}^T$ which requires computing the gradient of the kernel function for $m$ subsamples of chain history; and 2) computing the density of two multivariate normal distributions with non-diagonal covariance matrices. We aim to amortize these cost with the design of a new algorithm.

\begin{algorithm}[tb]
  \caption{Kernel Adaptive Metropolis-Hastings \cite{sejdinovic2014kernel}}
  \label{alg:KAM}
  \begin{algorithmic}
    \STATE \textbf{Input:} initial state $\theta_0 \in \Theta$, subsample size $m$, initial stepsize $\nu$, noise $\gamma$, stepsize adaptation $\epsilon$,  total iterations $T$, burin $B$,
    adaptation probabilities $\{p_t\}_{t=0}^\infty$,  kernel gradient $\nabla_{\theta _t} k(\theta_t,\theta')$, optimal acceptance $\alpha^*$
    
    \STATE \textbf{Let} $samples = \{ \}$ 
    \FOR{t in $0:T$}\vspace{0.3em}
    
        \STATE \textcolor{blue}{ $\triangleright$ \textbf{Propose next state by finding pre-image of a RKHS sample}}
        \STATE \textbf{Let} $\mathcal{M} = {\min(m, t+1)}$
        \STATE \textbf{with probability} $p_t$, random subsample $\mathbf{z} = \{z_i\}_{i=0}^{\mathcal{M} -1}$ from history $\{\theta_i\}_{i=0}^{t-1}$    
        \STATE \textbf{Let} $H = I - \frac{1}{\mathcal{M} }\mathbf{1}_{\mathcal{M}  \times \mathcal{M} }$ \text{the centering matrix in} $ \mathbb{R}^{\mathcal{M}  \times \mathcal{M} }$ 
        \STATE \textbf{Let} $M_{\mathbf{z},\theta_t} = 2[\nabla_{\theta _t}k(\theta_t,z_0),...,\nabla_{\theta_t}k(\theta_t,z_{m-1})]$ \text{the kernel gradient matrix in} $ \mathbb{R}^{d\times m}$ 
       \STATE  \textbf{Propose next state} $\theta'$ \text{through sampling from} $\mathcal{N}(\theta_t,\gamma^2I+\nu^2M_{\mathbf{z},\theta_t}HM_{\mathbf{z},\theta_t}^T)$
      
       \vspace{0.3em}
       \STATE \textcolor{blue}{ $\triangleright$ \textbf{Compute acceptance ratio $\alpha_t$}}
       \STATE \textbf{Compute multivariate normal PDF}, $q_{\mathbf{z}}(\theta'|\theta_t) = \mathcal{N}(\theta_t,\gamma^2I+\nu^2M_{\mathbf{z},\theta_t}HM_{\mathbf{z},\theta_t}^T)$ and $q_{\mathbf{z}}(\theta_t|\theta') = \mathcal{N}(\theta',\gamma^2I+\nu^2M_{\mathbf{z},\theta'}HM_{\mathbf{z},\theta'}^T)$.
       \STATE \textbf{Let} $\alpha_t = \min\left\{1, \frac{\uppi(\theta')q_{\mathbf{z}}(\theta_t|\theta')}{\uppi(\theta_t)q_{\mathbf{z}}(\theta'|\theta_t)}\right\}$
      
       \vspace{0.3em}
       \STATE \textcolor{blue}{ $\triangleright$ \textbf{Accept or reject the samples}}
       \STATE \textbf{With probability $\alpha_t$, $\theta_{t+1} = \theta'$}
      \STATE $samples = \{\theta_{t+1}, samples\}$ \textbf{If} {$t > B$} 
        
        \vspace{0.3em}
        
       \STATE \textcolor{blue}{ $\triangleright$ \textbf{Adapting the stepsize $\nu$}}
       \STATE $\nu = \exp(\log(\nu) + \eta[\alpha_t - \alpha^*])$ where $\eta = \frac{1}{(1+\epsilon)^t}$
      \ENDFOR
  \end{algorithmic}
\end{algorithm}

\subsection{Local vs. Global Covariance Estimate}
Standard AM and its variants estimate the empirical covariance matrix \cite{AdaMCMC_TM,Andrieu2008ATO,DIAM} base on the chain history of past samples $\{\theta_i\}_{i=0}^{t-1}$. However, these algorithms treat each sample equally, and updates the empirical covariance matrix regardless local landscape of posterior, therefore estimates a global covariance \cite{Roberts2007CouplingAE}.

KAM can adapt to the local covariance structure, because past samples are treated differently. KAM proposal computes the kernel gradient matrix $M_{\mathbf{z},\theta}$ on kernel functions $k(\theta,z_i)$, which will give larger values for $z_i$ closer to current position $\theta$ as long as the kernel is not a linear kernel. This proposal make better use of information encoded in past samples that are neighboring to current position even though samples are proposed from a Gaussian distribution.  
\subsection{Kernel Functions and Their Gradients}
Here we summarize kernels used in the original KAM paper, and their derivatives, which will be used to compute corresponding kernel gradient matrices.
\begin{itemize}
    \item \textbf{Linear Kernel.}  The kernel function is $k(\theta,z)=\theta^Tz$, its gradient w.r.t. $\theta $ is $\nabla_\theta k(\theta,z) = z$.\vspace{-0.2em}
    \item \textbf{Guassian (RBF) Kernel.} The kernel function is $k_{l}(\theta,z)=\exp{\left( -||\theta-z||_2^2/2l^2\right)}$, its gradient w.r.t. $\theta $ is $\nabla_\theta k_l(\theta,z) = k_l(\theta,z)(z-x)/l^2$.\vspace{-0.2em}
    \item \textbf{Matérn Kernel.} The kernel function is $k_{v,l}(\theta,z) =  \frac{1}{\Gamma(v)2^{v-1}}\left(\frac{\sqrt{2v}}{l}||\theta-z||_2 \right)^v K_v\left(\frac{\sqrt{2v}}{l}||\theta-z||_2 \right),$ where $K_v(\cdot)$ is a modified Bessel function of  the second kind for real order $v$, and $\Gamma(\cdot)$ is the gamma function. Its gradient w.r.t. $\theta $ is $\nabla_\theta k_{v,l}(\theta,z) = \frac{v}{l^2(v-1)}k_{v-1,l}(\theta,z)(z-x)$ \vspace{-0.2em}
\end{itemize}
\subsection{cKAM and cyclical stepsize}  \label{cKAM}
While KAM can adapt to the local covariance of a strongly non-linear target distribution, this behavior will hinder the sampler to properly explore a multimodal distribution. We want to design an algorithm that 1) unitizes KAM's ability to adapt to local covariance structure, while avoiding the large cost of making proposal in every iteration of the chain and 2) allows exploration to various modes of the target posterior \cite{multimodalAdaMCMC}. 

Inspired by recent success of cSG-MCMC \cite{zhang2020cyclical} in using cyclical step size for Bayesian Deep Learning, here we propose cyclical Kernel Adaptive Metropolis, cKAM, that splits the normal \textbf{Burnin} and \textbf{Sampling} phases in normal MCMC methods into \textit{cycles} of \textbf{Exploration} and \textbf{Sampling}.

\begin{algorithm}[tb]
  \caption{Cyclical Kernel Adaptive Metropolis}
  \label{alg:cKAM}
  \begin{algorithmic}

    \STATE \textbf{Input:} initial state $\theta_0 \in \Theta$, subsample size $m$, initial stepsize $\nu$, noise $\gamma$, stepsize adaptation $\epsilon$,  total iterations $T$, adaptation probabilities $\{p_t\}_{t=0}^\infty$,  kernel gradient $\nabla_{\theta _t} k(\theta_t,\theta')$, optimal acceptance $\alpha^*$,
    \STATE \textbf{new hyperparameters:} proporation of exploration $\beta$, number of cycles $M$
    
    \STATE \textbf{Let} $samples = \{ \}$ 
    \FOR{m in $0:M-1$}       
        \vspace{0.3em}
        \STATE \textcolor{blue}{ $\triangleright$ \textbf{Exploration phase}}
        \FOR{ $ 0 \leq \frac{\text{mod}{(t-1, \ceil{T/M})}}{\ceil{T/M}} \leq \beta$}
        \STATE \textbf{Same behavior as in KAM, except 1) no samples are added to $samples$, \\ 2) the random subsample $\mathbf{z}$ is only selected from chain history of current cycle}
        \ENDFOR
        
        \vspace{0.3em}

        \STATE \textcolor{blue}{ $\triangleright$ \textbf{Transition to Sampling phase, $\ceil{\frac{\text{mod}{(t-1, \ceil{T/M})}}{\ceil{T/M}}} = \beta$, $t=t_{exp}$}}
        
        \STATE \textbf{Let} $\nu_0 = ({2\nu_{exp}})/({\cos(\beta \uppi) + 1})$
        \STATE \textbf{Let} $\Sigma = (\frac{\gamma}{\nu_{exp}})^2I+ M_{\mathbf{z},\theta_t}HM_{\mathbf{z},\theta_t}^T$ 
        
        \vspace{0.3em}
        \STATE \textcolor{blue}{ $\triangleright$ \textbf{Sampling phase}}
        \FOR{ $ \beta < \frac{\text{mod}{(t-1, \ceil{T/M})}}{\ceil{T/M}} \leq 1$}
        \STATE $\nu_t = \frac{\nu_0}{2}\left[ \cos\left(\frac{\pi\ \text{mod}{(t-1, \ceil{T/M})}}{\ceil{T/M}}\right)+1\right]$
        \STATE \textbf{Propose $\theta'$ via sampling from $\mathcal{N}(\theta_t, \nu_t^2\Sigma)$}

       \STATE \textbf{With probability $\min \left\{1, \frac{\uppi(\theta')}{\uppi(\theta_t)}\right\}$, $\theta_{t+1} = \theta'$}
       \STATE $samples = \{\theta_{t+1}, samples\}$
        \ENDFOR
    \ENDFOR
  \end{algorithmic}
\end{algorithm}

\begin{figure}[H]
\centering
\includegraphics[width=0.45\textwidth]{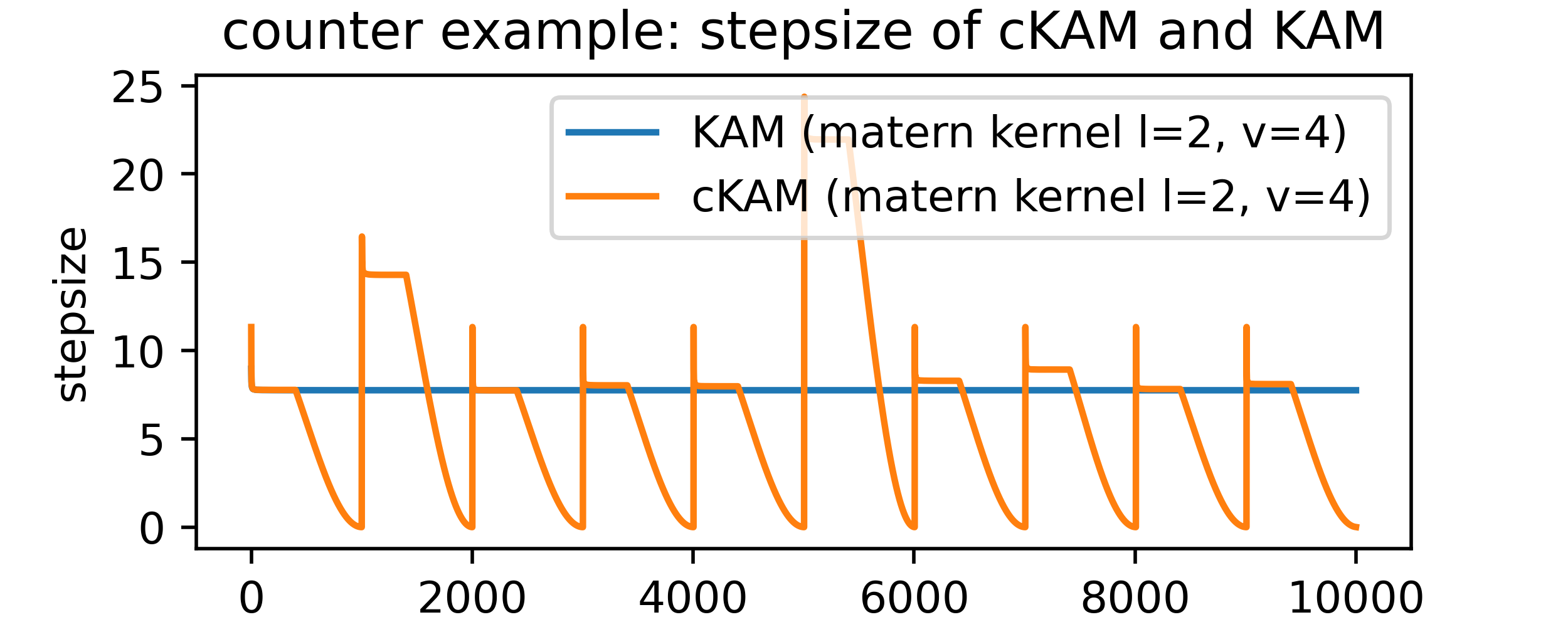}
\caption{\centering Stepsizes in the first 10 cycles of cKAM and corresponding 10000 iterations of KAM}
\label{fig:1}
\end{figure}

As in the cSG-MCMC paper, the cyclical cosine stepsize scheme is described as below: 
\[
    \nu_t = \frac{\nu_0}{2}\left[ \cos\left(\frac{\pi \ \text{mod}{(t-1, \ceil{T/M})}}{\ceil{T/M}}\right)+1\right],
\]
With $\nu_0$ being the initial stepsize, $T$ being the number of total iterations and $M$ being the number of cycles in the cSG-MCMC algorithm. However, since we need to adapt $\nu_t$ during the \textbf{Exploration} phase for better mixing, for $0 \leq \text{mod}{(t-1, \ceil{T/M})} \leq \beta \ceil{T/M}$, we will use the recurrent relation:
\[
    \log(\nu_{t+1}) = \log(\nu_{t}) + \eta [\alpha(\theta_{t}, x_{t-1}) - \alpha^*], 
\]
where $\alpha^*$ is the optimal acceptance ratio \cite{0.234Optimal} usual to be 0.234, $\beta$ is the proportion of exploration phase, and $\eta$ is the `stepsize' \footnote{ `stepsize' here refers to the magnitude of a step of random walk} in the Robbins–Monro algorithm. Therefore, we use a modified cyclical stepsize scheme that computes the theoretical initial stepsize $\nu_0^m$ for cycle $m$ from the last stepsize in \textbf{Exploration}, which leads the formula for $\nu_0^m$ to be:
\[
    \nu_0^m = \frac{2\nu_{exp}^m}{\cos(\beta \pi) + 1}
\]
This leads to cKAM algorithm, as in Algorithm \ref{alg:cKAM}. We show stepsizes in the first 10 cycles of cKAM for the Bimodal experiment \ref{bimodal} in figure \ref{fig:1}.

\subsection{Exploration and Sampling in cKAM}
\subsubsection*{\textbf{Exploration Phase}\label{sec:exploration}} 
    
At \textbf{Exploration} of a cycle, normal KAM proposal is used to propose a next state. With the noise term $\gamma$, initial iterations of this cycle will explore more. We emphasize that this property is made possible by selecting the random subsample $\mathbf{z}$ from the chain history of current cycle. If otherwise, and $\mathbf{z}$ can be selected from the entire chain history, the property of exploration will be lost as the proposed state will depend on the positions in previous cycles close to the current position. \par
    
However, we admit that the initial positions in the current cycle would depend on the very last sample in the previous cycle, just like the initial positions in the chain of any other MCMC algorithms would depend on the initial state $\theta_0$. One can follow the same analogy made in the cSG-MCMC paper, and think of \textbf{Exploration} as a warm restart to explore a different part of posterior following local covariance structure. By using adaptive stepsizes in this phase, cKAM can experience enough perturbation to escape current mode, and can benefit from a better mixing rate by optimizing for a target acceptance ratio $\alpha^*$.

\subsubsection*{\textbf{Sampling Phase}\label{sec:sampling}}
    
The stepsizes in this phase follows the pattern described in \ref{cKAM}, with a theoretical $\nu_0$ computed from the final stepsize in \textbf{Exploration} as if \textbf{Exploration} also used a cyclical stepsize schedule. The stepsize in \textbf{Sampling} decreases rapidly, allow cKAM to better explore the landscape of local regions. \par 
    
In order to greatly reduce computational cost of making a proposal, we unconventionally change the proposal from KAM proposal to a RW proposal. Although the computation cost of KAM proposal might be still much smaller than evaluation of likelihood functions when using vanilla MH, this reduction will become significant if mini-batched MH algorithms are used, such as TunaMH \cite{tunaMH}, FlyMC \cite{flyMC} and others. 

\subsection{Ergodicity of cKAM}
Here we sketch out a rough proof based on previous works, to show that cKAM samples from the exact posterior distribution. 
\par First, we recognize that cKAM belongs to the framework proposed by Tierney \cite{Tierney1994MarkovCF} of MCMC mixture using cycles, where proposals $q_1,...,q_k$, when used individually can lead to the correct target posterior, are used in return in a cycle. Indeed, with a cycle, except for one position collected in the transition from \textbf{Exploration} to \textbf{Sampling}, all others are from draw from either KAM or RW proposals that has been applied with either MH or standard Metropolis. Furthermore, it is obvious to see the irreducibility and aperiodic requirements for the mixture \cite{Tierney1994MarkovCF} are satisfied, since KAM proposal is asymmetric. cKAM doesn't need to show weak convergence in Theorem 2 of \cite{zhang2020cyclical} which is intended to show asymptotic bound on the bias and MSE caused by the absence of Metropolis-Hastings in SGLD \cite{sgld}. \par
    
Second, we need to justify for the sudden change in the type of proposal (KAM proposal $\rightarrow$ RW proposal). This change in the type of proposal is seen in algorithms that uses Delayed Rejection \cite{DR1,DR2,DRAM} where conventionally cheaper-to-compute proposals are tried first and upon rejection, more sophisticated proposals are used. Here we can think of the transition as the KAM proposal taking a first try and always fail, with RW proposal making the second move. Note that if we want to explicitly encode this in our algorithm as in Equation 3 of \cite{DRAM} and ensure reversibility, we need to have a position $\theta'$ proposed from KAM proposal $q_{\mathbf{z}}(\theta'|\theta_{exp})$, forcefully reject it, and propose a second position $\theta''$ through random walk $q_{RW}(\theta''|\theta')$, and accepts it with acceptance probability
\[
\min\left\{1, \frac{\uppi(\theta'')q_{\mathbf{z}}(\theta_{exp}|\theta')q_{RW}(\theta_{exp}|\theta')[1-\alpha(\theta'|\theta'')]}{\uppi(\theta_{exp})q_{\mathbf{z}}(\theta'|\theta_{exp})q_{RW}(\theta''|\theta')[1-\alpha(\theta'|\theta_{exp})] }\right\}
\]

We decide to ignore this, since the sample proposed would be exactly the same just with a different acceptance ratio; this is not the case that the sampler suddenly proposes a very distant next position. Any bias introduced from this step will be smoothed out by the rest of the positions in the cycle. We conclude that the cKAM sampler is ergodic.\par

Furthermore, we are only collecting samples in \textbf{Sampling}, which only used the RW proposal for all the samples. Because of this, one can think of \textbf{Exploration} as finding a good initial state for a RW sampler in each cycle. Rather than continuing adapting a covariance matrix, we take the term $(\gamma^2I+\nu^2M_{\mathbf{z},\theta_t}HM_{\mathbf{z},\theta_t}^T$ as $\nu^2\Sigma)$ to be a fuzzy estimate of local covariance matrix $\Sigma$ for RW proposals. 

\section{Experiments}
We fix our comparison to the following samplers:
\begin{itemize}[noitemsep, topsep=0pt, partopsep=0pt]
    \item \textbf{(RW)} Random-Walk  with isotropic proposal $q(\cdot | \theta_t) = \mathcal{N}({\cdot | \theta_t, \nu^2I})$ where $\nu$ is the stepsize 
    \item \textbf{(AM)} Adaptive Metropolis  with a learned covariance matrix and fixed stepsize (Algorithm 2 in \cite{Andrieu2008ATO}) 
    \item \textbf{(RBAM)} Rao-Blackwellised AM algorithm using the ``average position" between $\theta_t$ and $\theta'|\theta_t$ (Algorithm 3 in \cite{Andrieu2008ATO}) 
    \item \textbf{(GAM)} Adaptive Metropolis with a learned covariance matrix and a global stepsize adapting to the optimal acceptance $\alpha^*=0.234$ (Algorithm 4 in \cite{Andrieu2008ATO}) 
    \item \textbf{(KAM)} Kernel Adaptive Metropolis-Hastings  
    \item \textbf{(cKAM)} Our cyclical Kernel Adaptive Metropolis 
\end{itemize}
We also implement decaying noise schedule $\gamma_t = a(b+t)^\text{-decay\_rate}$ for KAM, which is set to a constant of $\gamma=0.2$ in the original KAM paper. 

\subsection{Difficult 2D Bimodal distribution}
\label{bimodal}
\begin{figure} [H]
\centering
\includegraphics[width=0.4\textwidth]{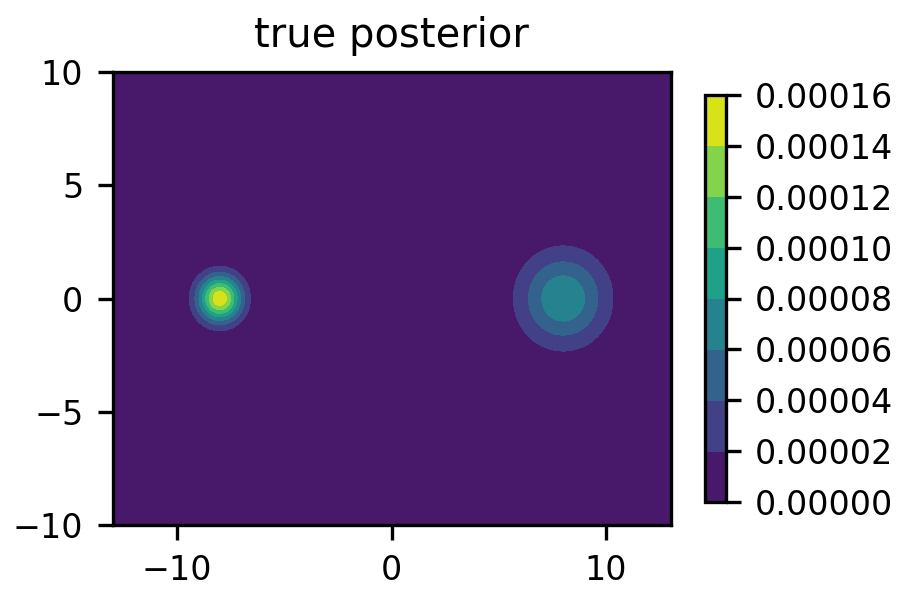}  
\caption{\centering Density plot for the 2d bimodal distribution}
\label{fig:true posterior}
\end{figure}

To demonstrate the importance of exploration, we here provide a synthetic distribution. We want to emphasize that this is a harder synthetic task than 1) double well experiment used in \cite{double_well_1,double_well_2, zhang2020amagold} and 2) two-dimensional nonlinear distributions as described in \cite{sejdinovic2014kernel,yin2018semiimplicit,zhang2020amagold}, since our proposed task is a two-dimensional distribution of two more distant modes. 
The density plot of the true distribution is show in figure \ref{fig:true posterior}. The analytical form of the distribution and how we set our hyperparameters are described in appendix \ref{bimodal_appendix}. 
 \vspace{-0.5em}
We plot symmetric Kullback–Leibler (KL) divergence versus wall-clock time after any sample is collected. As can be seen in figure \ref{fig:counter} and in appendix \ref{bimodal_appendix}, all the adaptive MCMC samplers only converges to one mode. The well-tuned RW (using an extremely unreasonable large $\nu = \sqrt{15}$ as oppose to normal value $2.38/\sqrt{2}$) seems to be asymptotically converging, but is doing so slower than cKAM. While KAM seems to able to escape initialization of $[-8, 0]$ due to large initial noise $\gamma_0$, it gets trapped in the other mode around $[8, 0]$. In contrast, cKAM is able to find and fully explore both modes, regardless of the initial position. cKAM leverages large stepsizes in \textbf{Exploration} to find other modes, and exploits a local mode in \textbf{Sampling} when having smaller stepsizes.   
 
\begin{figure}[H]
\centering
\includegraphics[width=0.45\textwidth]{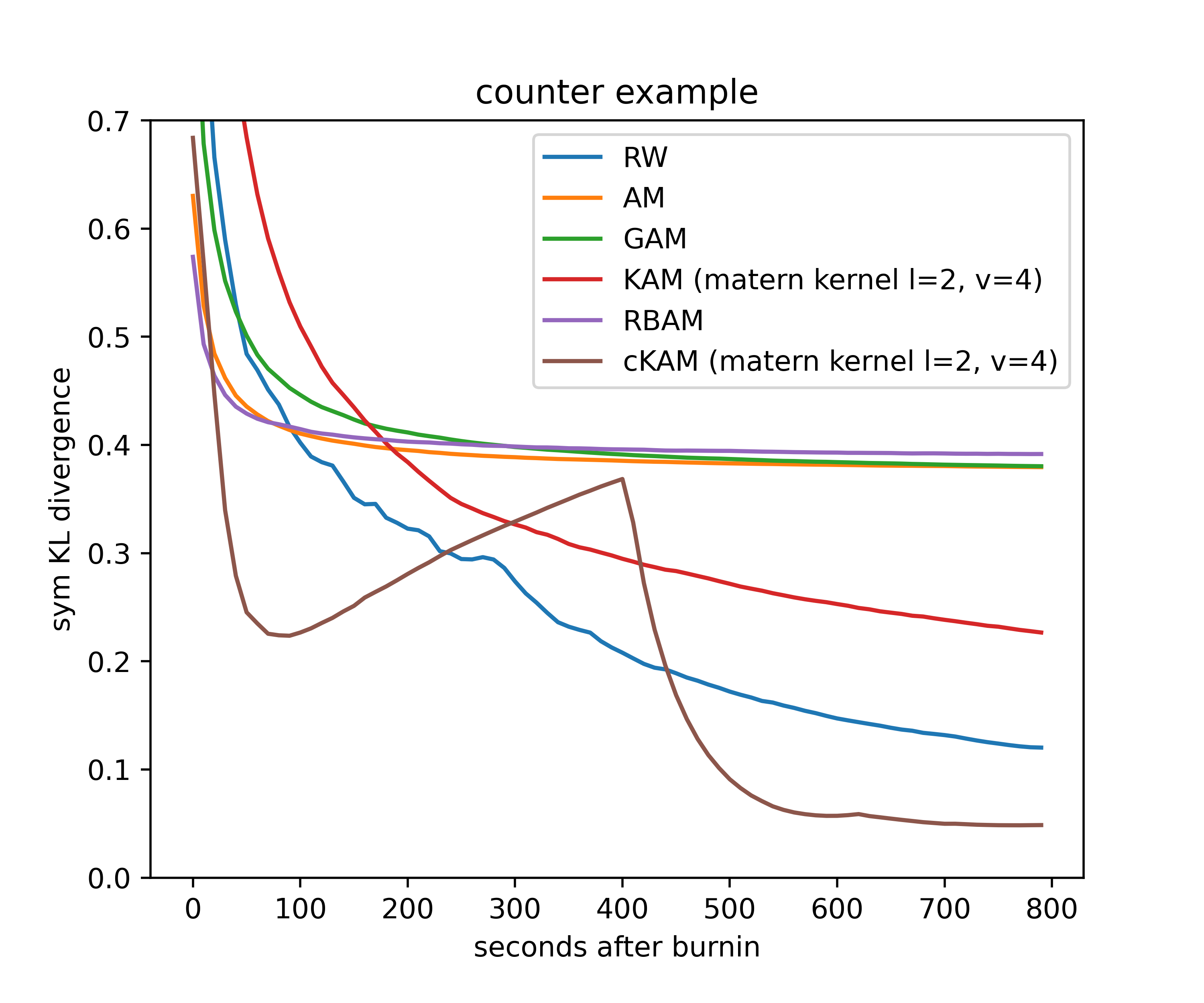}
\vspace{-1em}\caption{KL divergence against time for all samplers}
\end{figure}   

We continued running cKAM for more time, and it converges to the true posterior as shown below:
\begin{figure}[H]
\centering
\includegraphics[width=0.4\textwidth]{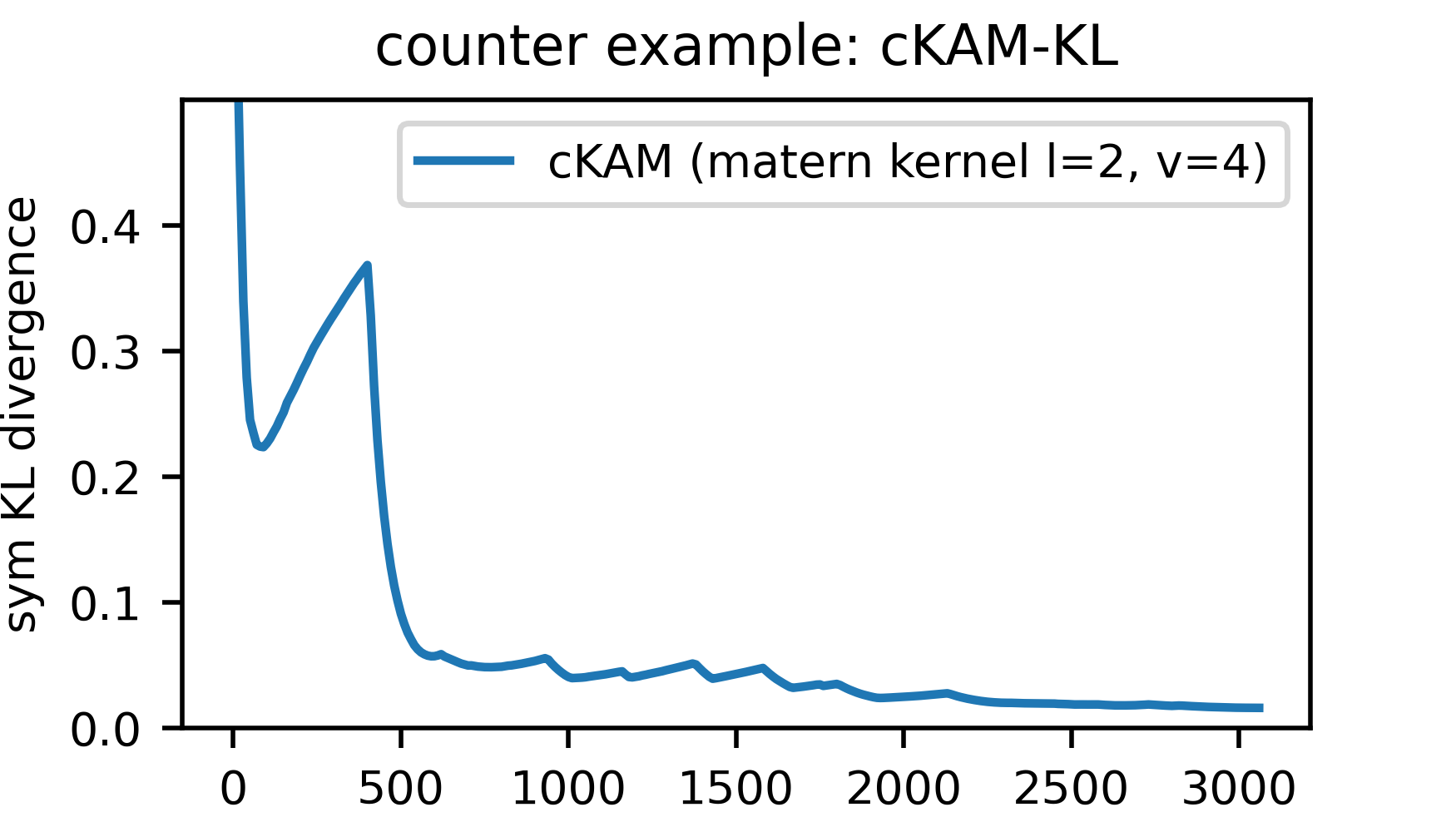}
 \vspace{-1em}
\caption{KL divergence against time for cKAM}
\label{fig:KL cKAM}
\end{figure}
 
Figure \ref{fig:KL cKAM} gives us some intuition behind cKAM's behavior: around time 100 second, cKAM jumps to another mode which explains the increase in KL divergence. cKAM roughly covers both modes at time 500 second, then it jumps between modes in \textbf{Exploration} and explore the local landscape in \textbf{Sampling}, which explains the small oscillations in the figure.  We also show the effect of using different kernels for both KAM and cKAM in appendix \ref{bimodal_appendix}.

\newpage
\subsection{2D Gaussian Mixtures with increasing Std}
We then proceed with a more realistic setting, where 5 Gaussian distributions with increasing standard deviation. This should be a easier task than the Bimodal one since all individual Gaussians are close to each other. In figure \ref{fig:std_posterior} we show its true density plot. The analytical form is given in appendix \ref{std_appendix}.  \vspace{-0.5em}
\begin{figure}[H]
\centering
\includegraphics[width=0.4\textwidth]{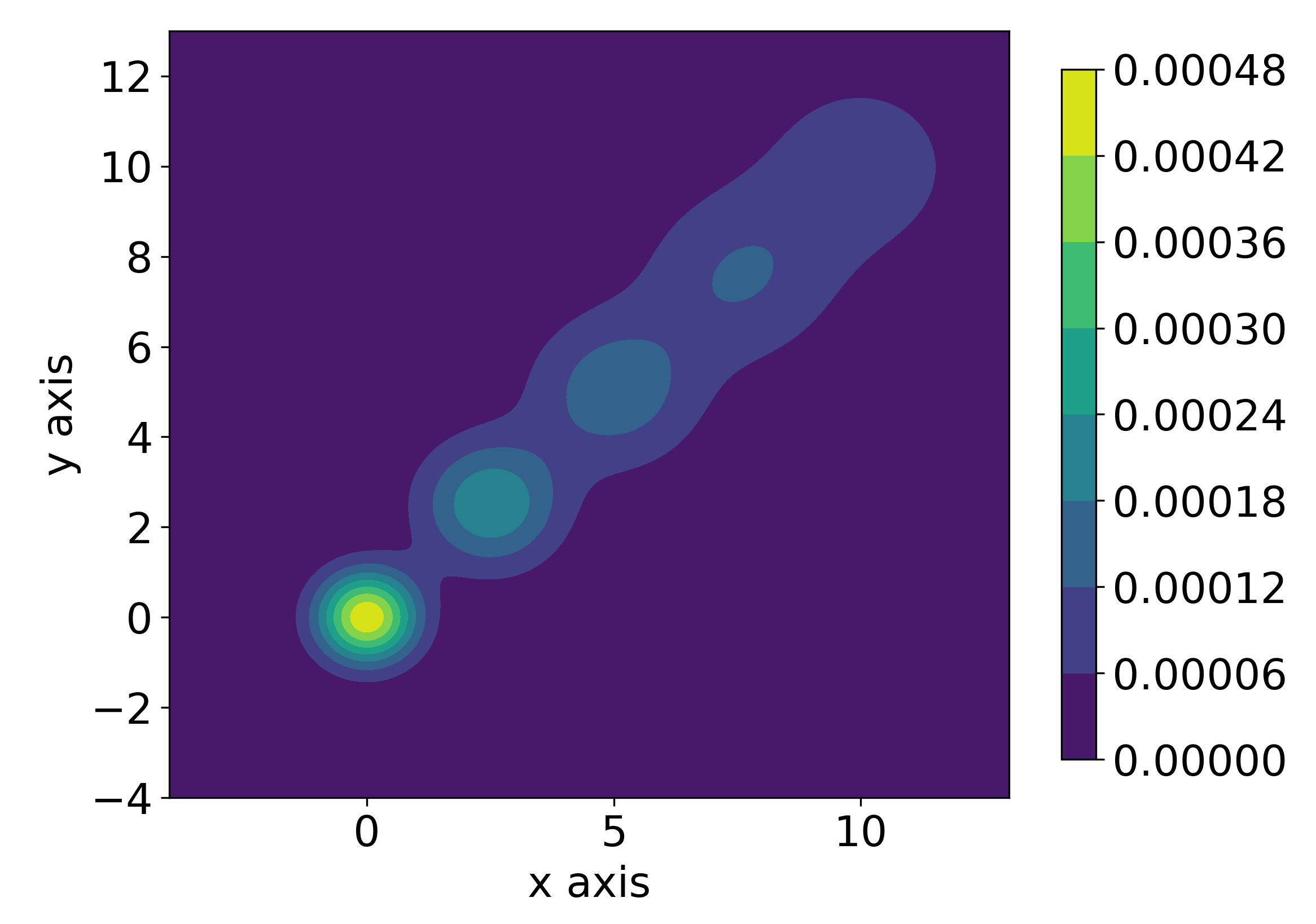}\vspace{-2em}
\caption{\centering  Density plot for the 2d Gaussian Mixtures}
 
\label{fig:std_posterior}
\end{figure} 

We show the KL divergence versus wall-clock time after any sample is collected. 
\begin{figure}[H]
\centering
\includegraphics[width=0.5\textwidth]{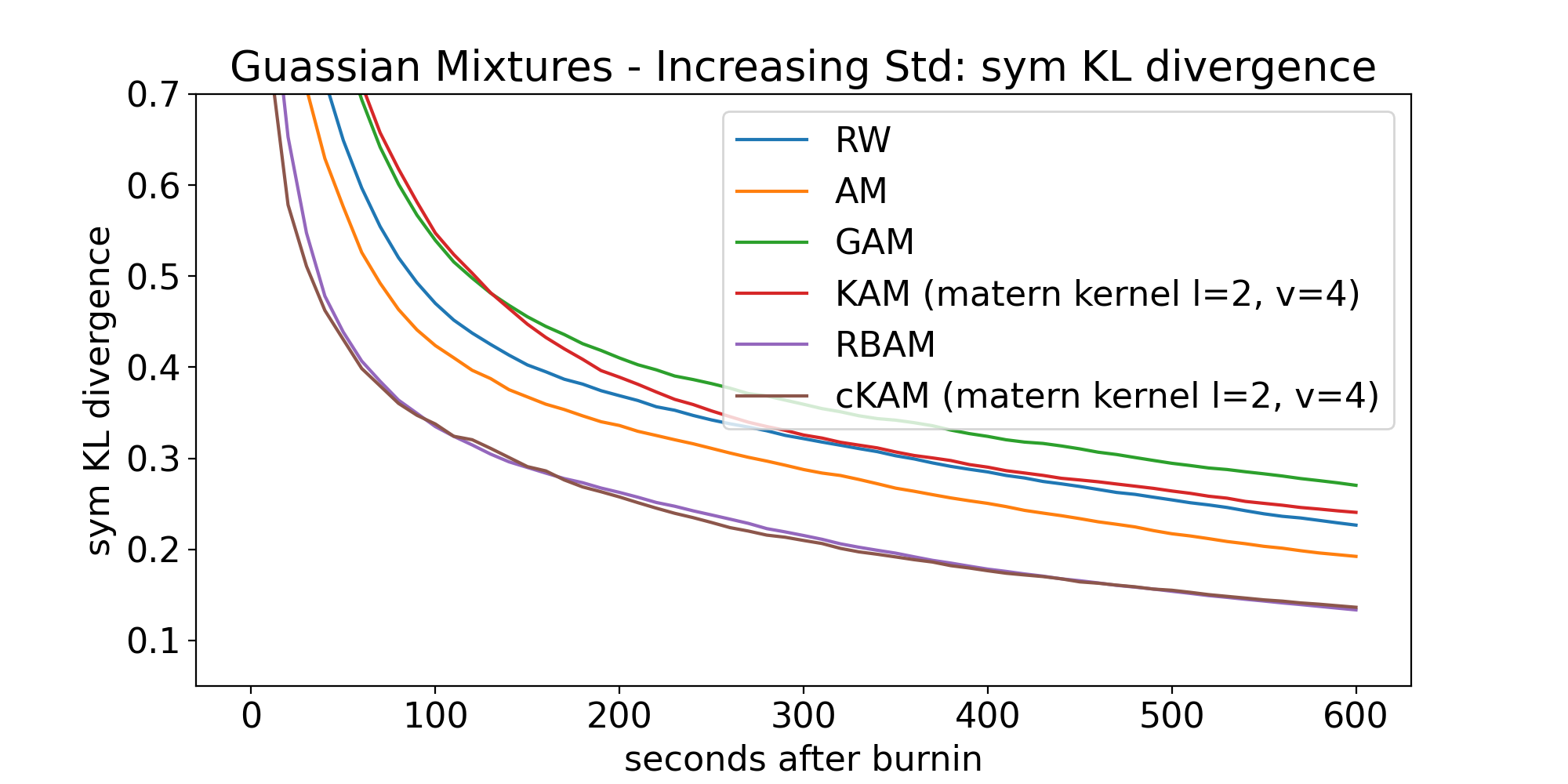}
\caption{\centering KL divergence against time for all samplers}
\label{fig:std_posterior_all}
\end{figure} 
 
As can be seen, while all the samplers seem to be asymptotically converging, cKAM is converging faster than KAM, while on par with the best adaptive method RBAM. After running cKAM with more time budget, it converges to the true posterior as show in Figure \ref{fig:std_posterior}

\begin{figure}[H]
\centering
\includegraphics[width=0.4\textwidth]{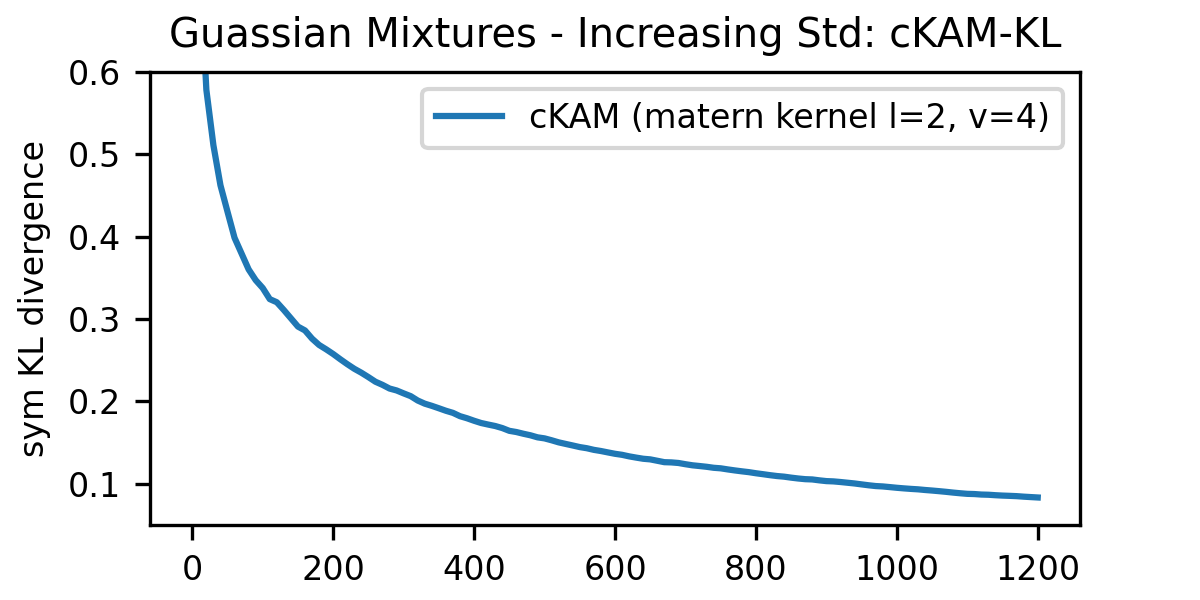}  
\caption{\centering KL divergence against time for cKAM} 
\label{fig:std_posterior}
\end{figure} 

\subsection{High-dimensional Gaussian Mixtures}
We construct a 32-dimensional Gaussian Mixtures with $5^{32}$ 32-dimensional Gaussian distributions with different mean and same covariance arranged in a hyper-cube. We plot the marginalized density over dimension 2 and 7. 
\begin{figure}[H]
\centering
\includegraphics[width=0.4\textwidth]{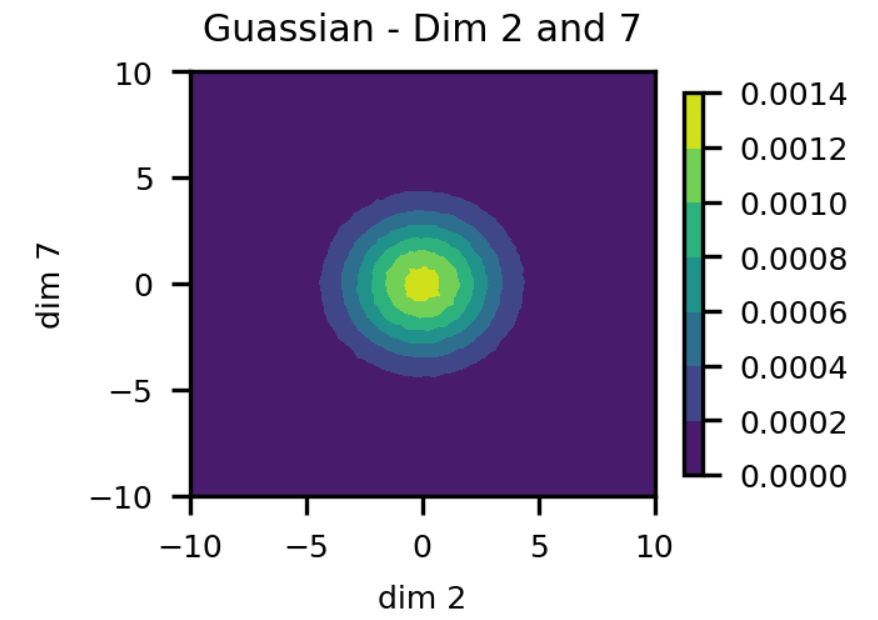}
\centering
\caption{\centering Dimensions 2 and 7 of the high-dimensional Gaussian Mixtures}
\label{fig:high-dim-2-7}
\end{figure} 

We show the KL divergence versus wall-clock time after any sample is collected. Due to the curse of dimensionality, it is impossible to generate exact true posterior distribution since it is computationally intractable to build a meshgrid in high-dimensional space (even it the most smallest grid that has 2 slots for each dimension will take $2^{32} * 4\text{B} = 4 \text{GiB}$ memory in total). Since our true posterior distribution is a Gaussian distribution that we know its analytical form in any dimension, we can compute the theoretical marginal posterior probability in one dimension and then compute the symmetric KL divergence with the samples in such dimension. We finally average our results and gain the mean symmetric KL-divergence for such sample.
 
\begin{figure}[H]
\centering
\includegraphics[width=0.5\textwidth]{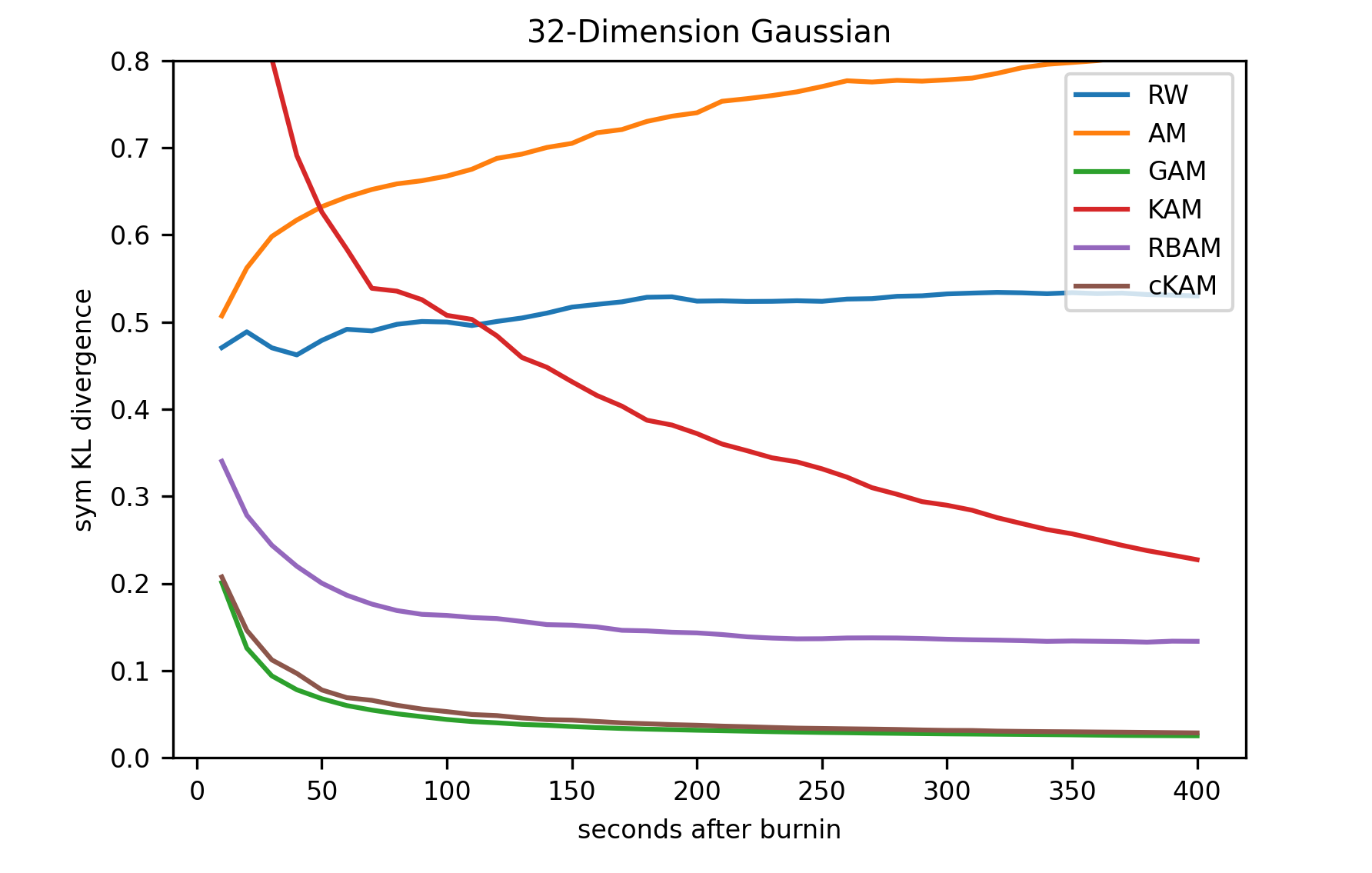}
\vspace{-2em}
\caption{\centering KL divergence against time for all samplers}
\end{figure} 

As can be seen, cKAM outperforms most samplers, with similar performance as GAM. We also observe that during our experiments, the performance of RW proposal was not stable and sometimes it might not converge to the true posterior easily (we needed to tune the initial weights to make RW proposal converge).  We guess this might be due to the curse of dimensionality. AM's KL divergence increases, possibly due to the curse of dimensionality, as the AM sampler explores the posterior poorly, leading to unstable results.


\section{Conclusion}
We design the cyclical Kernel Adaptive Metropolis (cKAM) algorithm, a variant of the Kernel Adaptive Metropolis-Hastings sampler that allows better exploration of the posterior distribution, while still retaining the local covariance adaptation property. We empirically show that standard adaptive MCMC samplers that continuously conform to either the local or global covariance structure will be incapable of effectively explore multimodal distributions. In experiment, cKAM performs better or on par with well-tuned adaptive samplers, suggesting that our trick of using a fuzzy estimate of local covariance matrix to greatly save computation cost works well in practice. Further extension could include a theoretical convergence analysis on cKAM, and an ablation experiment to demonstrate the importance of using our method compared with proposals that estimate the global covariance, under the cyclical stepsize scheme.

\newpage

\bibliography{ref}
\bibliographystyle{icml2021}

\onecolumn
\appendix

\section{EXPERIMENTAL RESULTS}
\subsection{Difficult 2D Bimodal distribution}
\label{bimodal_appendix}
We design the two-dimensional distribution to have two modes, each relatively distant from each other. The analytical form of the distribution is:
\[
    \mathcal{P}(x) = \frac{1}{2}\mathcal{N}\left(x|\mu_1, \Sigma_1 \right) + \frac{1}{2}\mathcal{N}\left(x|\mu_2, \Sigma_2\right)
\]

Where $\mu_1 = \begin{bmatrix}-8\\0\end{bmatrix}, \Sigma_1 = \begin{bmatrix}0.5& 0\\0&0.5\end{bmatrix} $ and $\mu_2 = \begin{bmatrix}8\\0\end{bmatrix}, \Sigma_2 = \begin{bmatrix}2& 0\\0&2\end{bmatrix} $. \vspace{2mm}
 
We initialize all the samplers from $\mu_1$, use a subsample size of 30 for KAM and 50 for cKAM, and run following hyperparameter settings for our samplers:\vspace{-4mm}
\begin{table}[H]
  \caption{Hyperparameter settings for Bimodal distribution}
  \label{tab:hyp_bimodal}
  \centering
  \begin{tabular}{lcccccc}
    \toprule      
       & (initial) stepsize $\nu$ & RM stepsize $\eta$ & RM rate $\epsilon$   & $T/M$ &$\beta$ &optimal acceptance $\alpha^*$\\
    \midrule
    \vspace{.1cm}
    RW & $\sqrt{15}$ & N/A & N/A &N/A&N/A& N/A  \\
    \vspace{.1cm}
   AM & $\frac{2.38}{\sqrt{2}}$ & 0.1 & N/A &N/A& N/A  &N/A\\
    \vspace{.1cm}
    RBAM & N/A & 0.1 & N/A &N/A& N/A &N/A \\
    \vspace{.1cm}
    GAM &$\frac{2.38}{\sqrt{2}}$ & N/A & 0.75 & N/A &N/A  &0.234 \\
   KAM & $\frac{2*2.38}{\sqrt{2}}$ & N/A & 0.75 & N/A& N/A  &0.234 \\
    cKAM & $\frac{2*2.38}{\sqrt{2}}$ & N/A & 0.75 & 1000 & 0.4 &0.234 \\
    \bottomrule
  \end{tabular}
\end{table}
 Where RM refers to Robbins–Monro optimization, where $\eta$ is not the stepsize for proposal, but for adaptation of sampler parameters. $\epsilon$ is vanishing rate of adaptation for $\eta_t = (1+t)^\epsilon$, used in GAM, KAM and cKAM. RBAM does not need an initial stepsize, but an initial $\Sigma_0$, which we set to $I$. We allow KAM and cKAM to have a large initial stepsize of $\frac{2*2.38}{\sqrt{2}}$ to encourage exploration, but because the stepsize is adaptive, as in the case of GAM, we still can have better mixing by optimizing stepsize to have running acceptance ratio $\alpha$ to be close to $\alpha^*$. Since we are running our samplers against time, we here do not provide a fixed number of iterations for samplers, but provide $T/M$ which is number of iterations per cycle for cKAM. We use the Matérn Kernel with $l = 2$ and $v=4$ for cKAM and KAM when comparing with other methods. Below is the marginal probability of all the samplers, and the true distribution. 
\begin{figure}[H]
\centering
\includegraphics[width=1\textwidth]{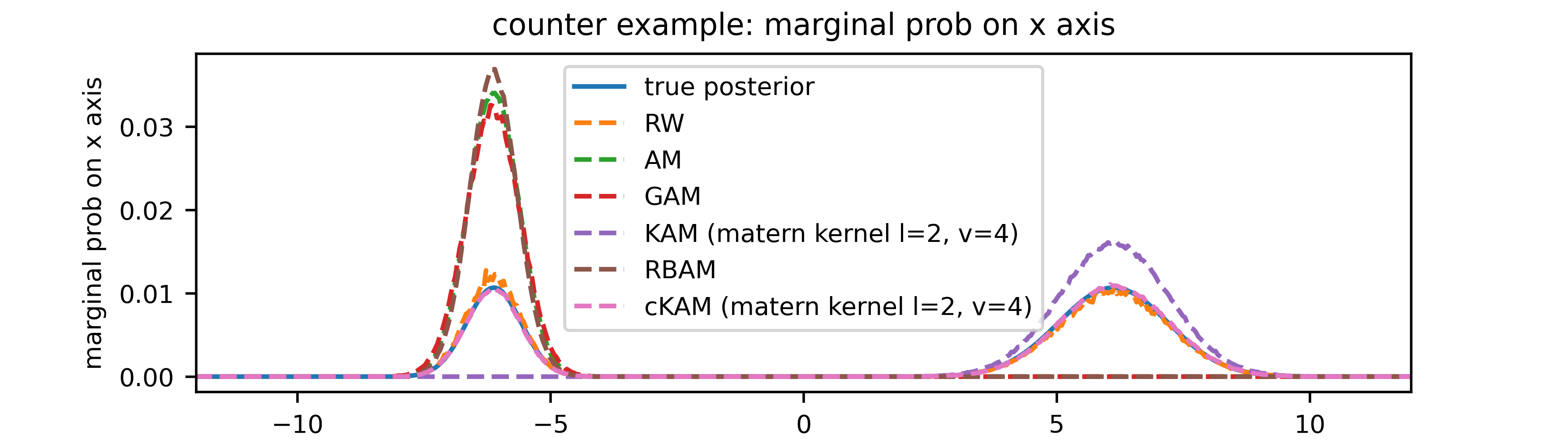} 
\caption{\centering Marginal distribution on the first dimension for all samplers}
\label{fig:counter}
\end{figure}

We plotted the Effective Sample Size against time for all the samplers:
\begin{figure}[H]
\centering
\includegraphics[width=0.65\textwidth]{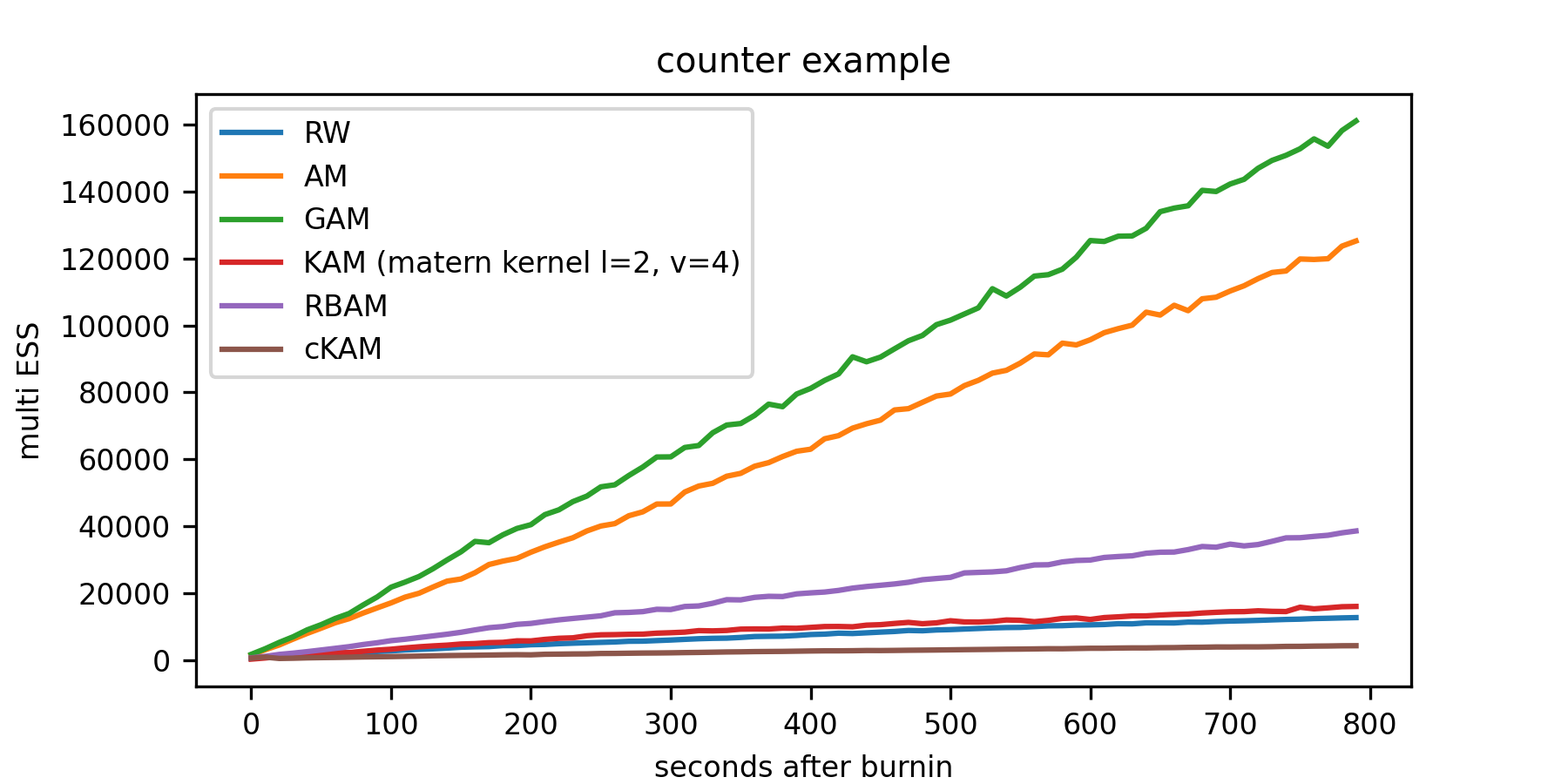} 
\caption{\centering ESS against time}
\end{figure} 
Although cKAM has the lowest ESS per time, this is because all other samplers excepts RW converge to the wrong distribution. We also show all the density plot for the samples collected by different samplers:
\begin{figure}[H]
\centering
\includegraphics[width=0.4\textwidth]{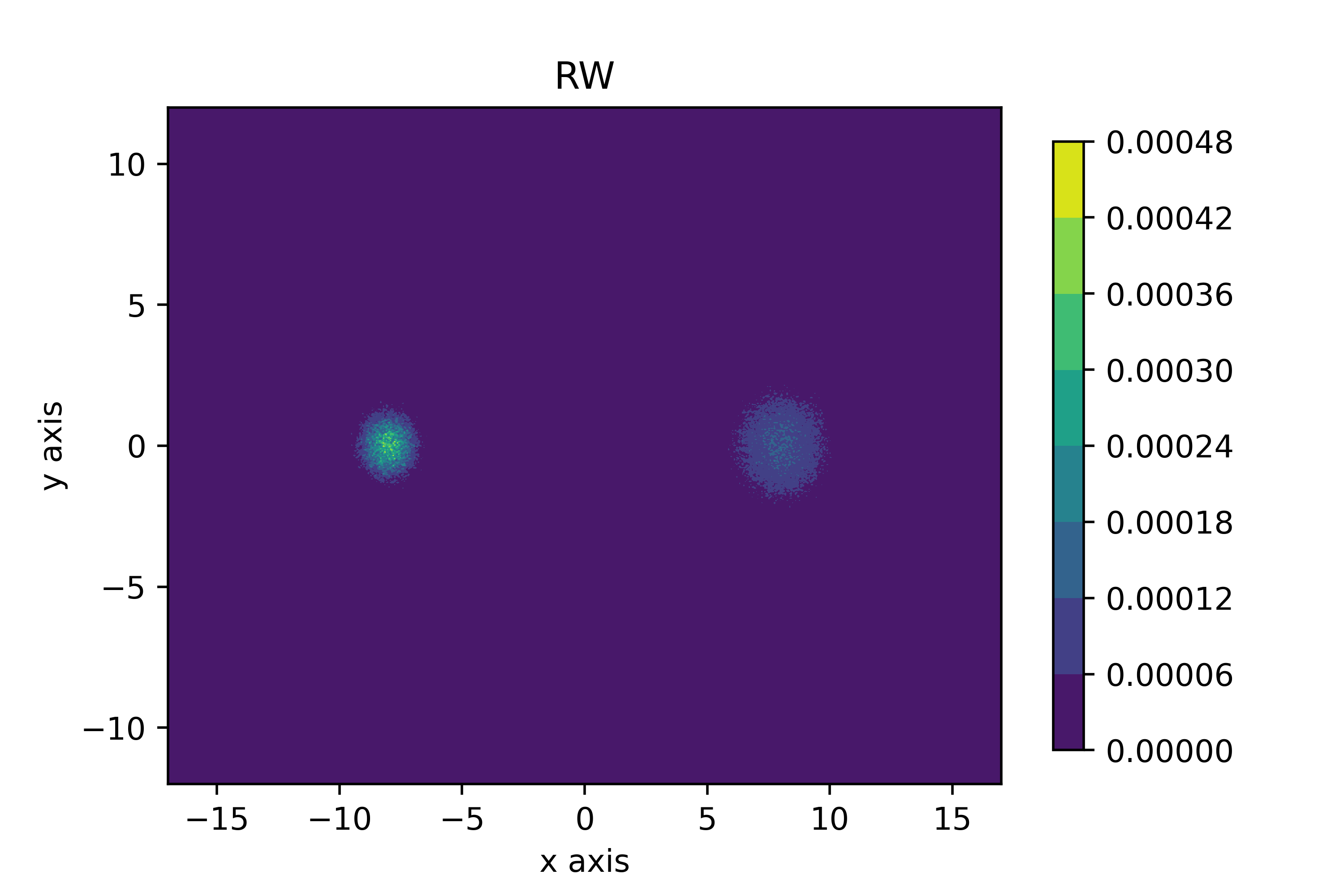}~ 
\includegraphics[width=0.4\textwidth]{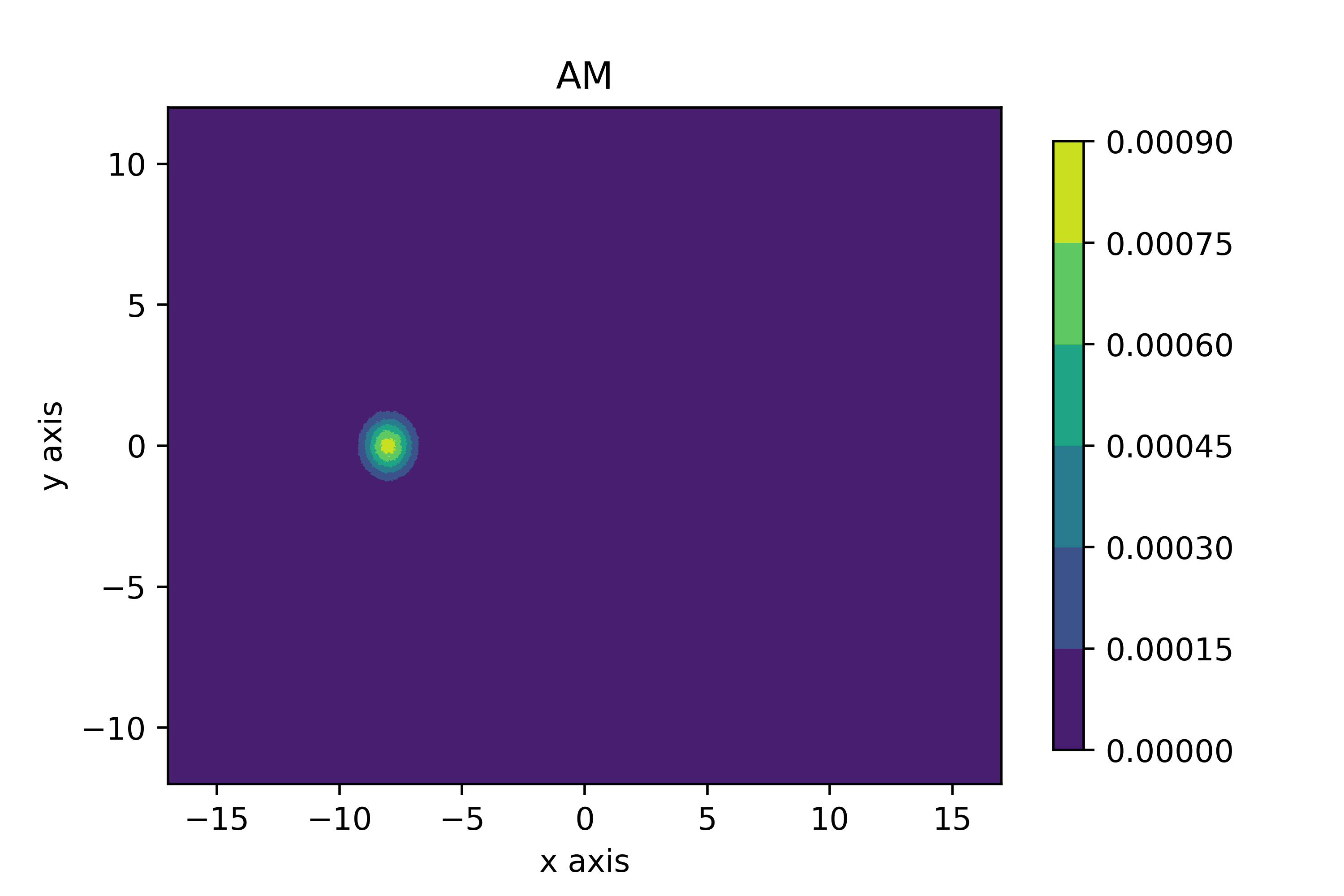}\\ 
\includegraphics[width=0.4\textwidth]{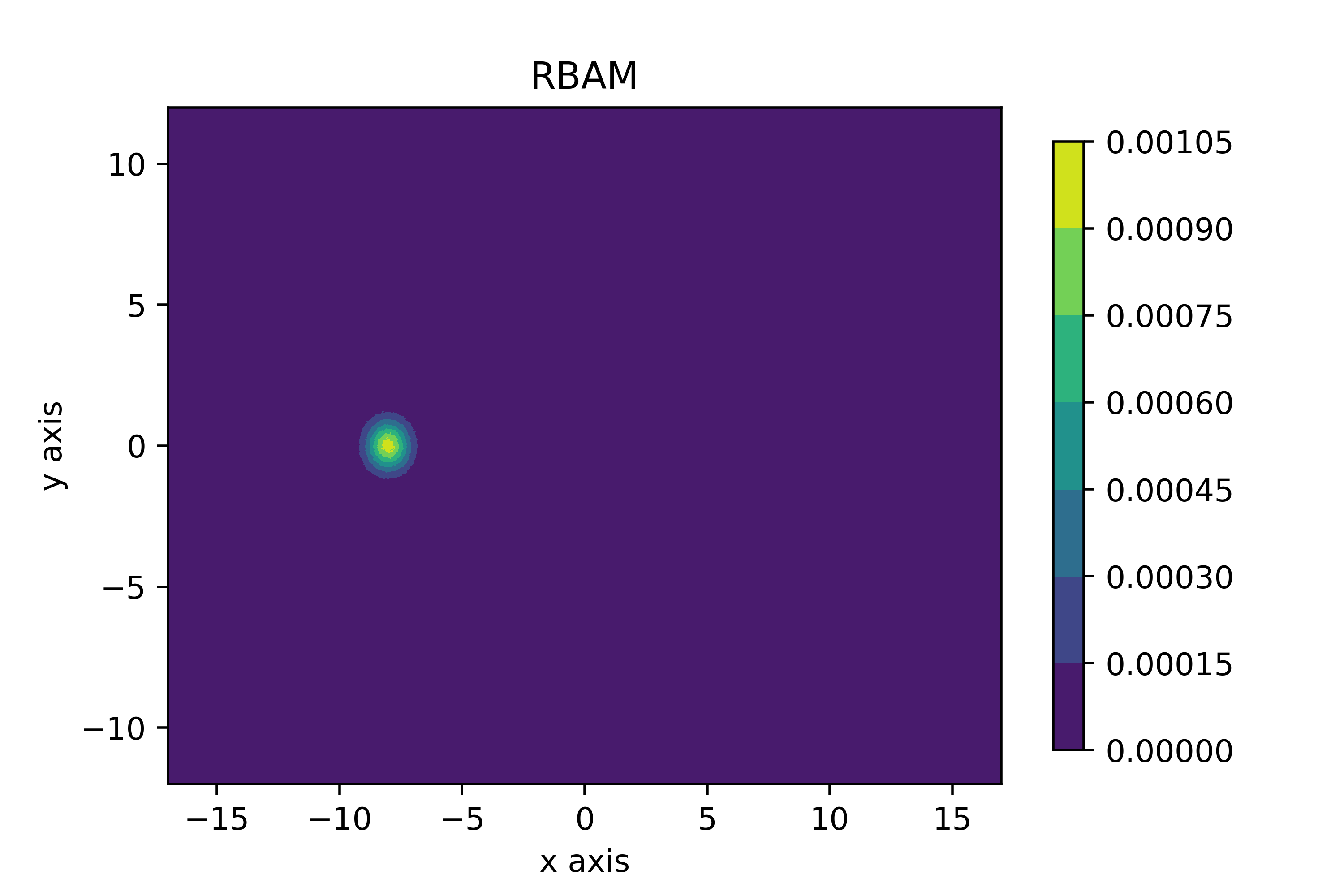}~
\includegraphics[width=0.4\textwidth]{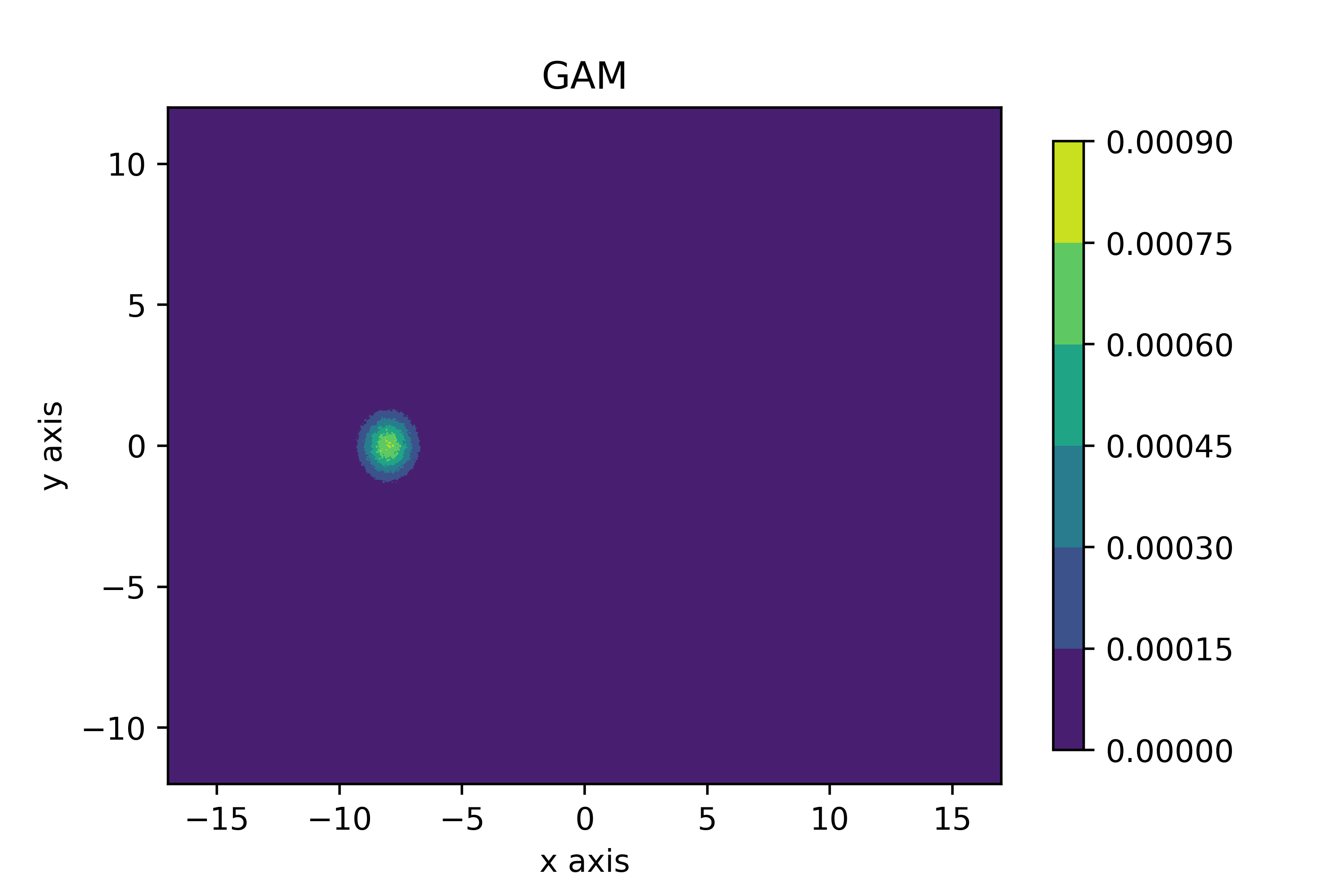}\\
\includegraphics[width=0.4\textwidth]{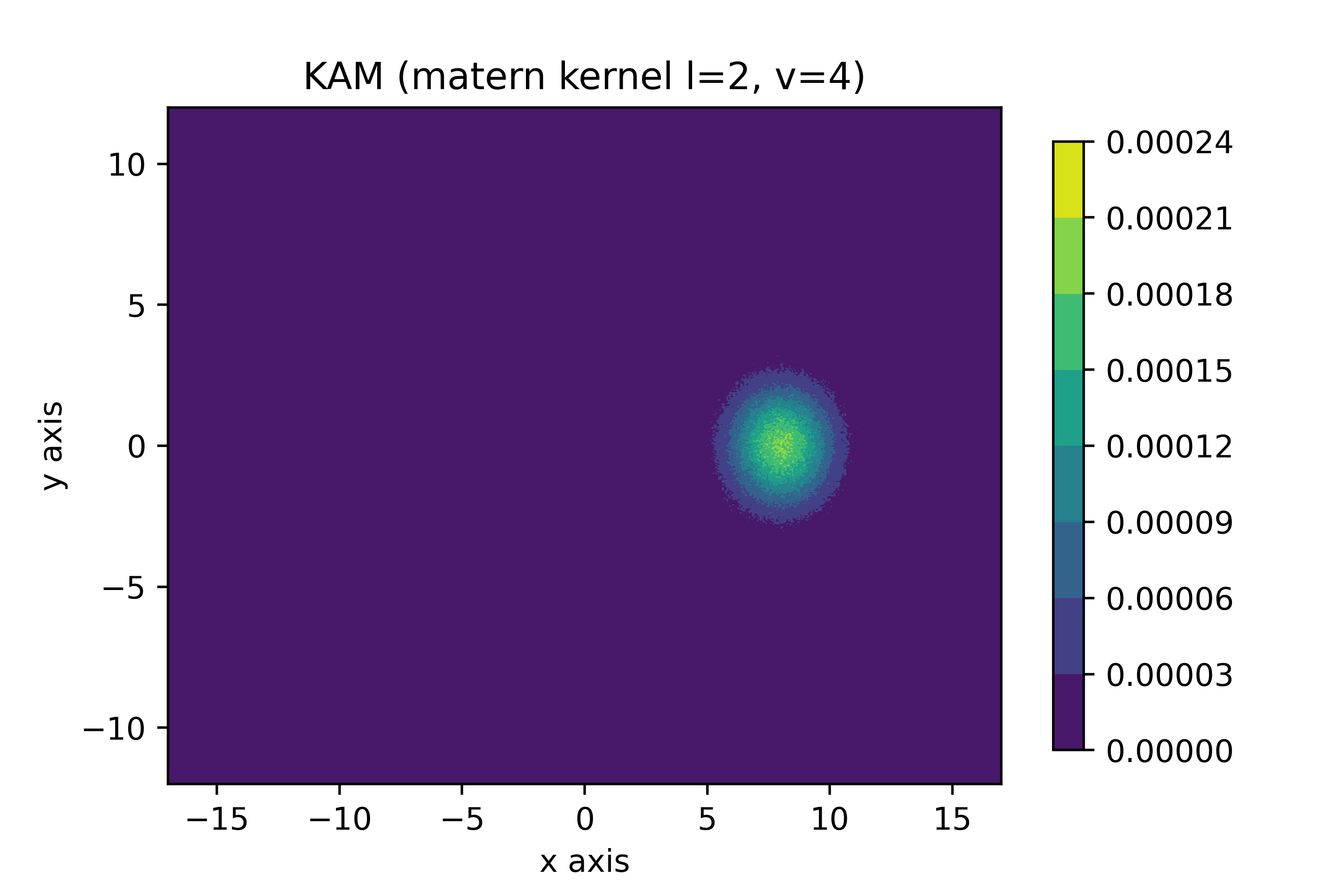}~ 
\includegraphics[width=0.4\textwidth]{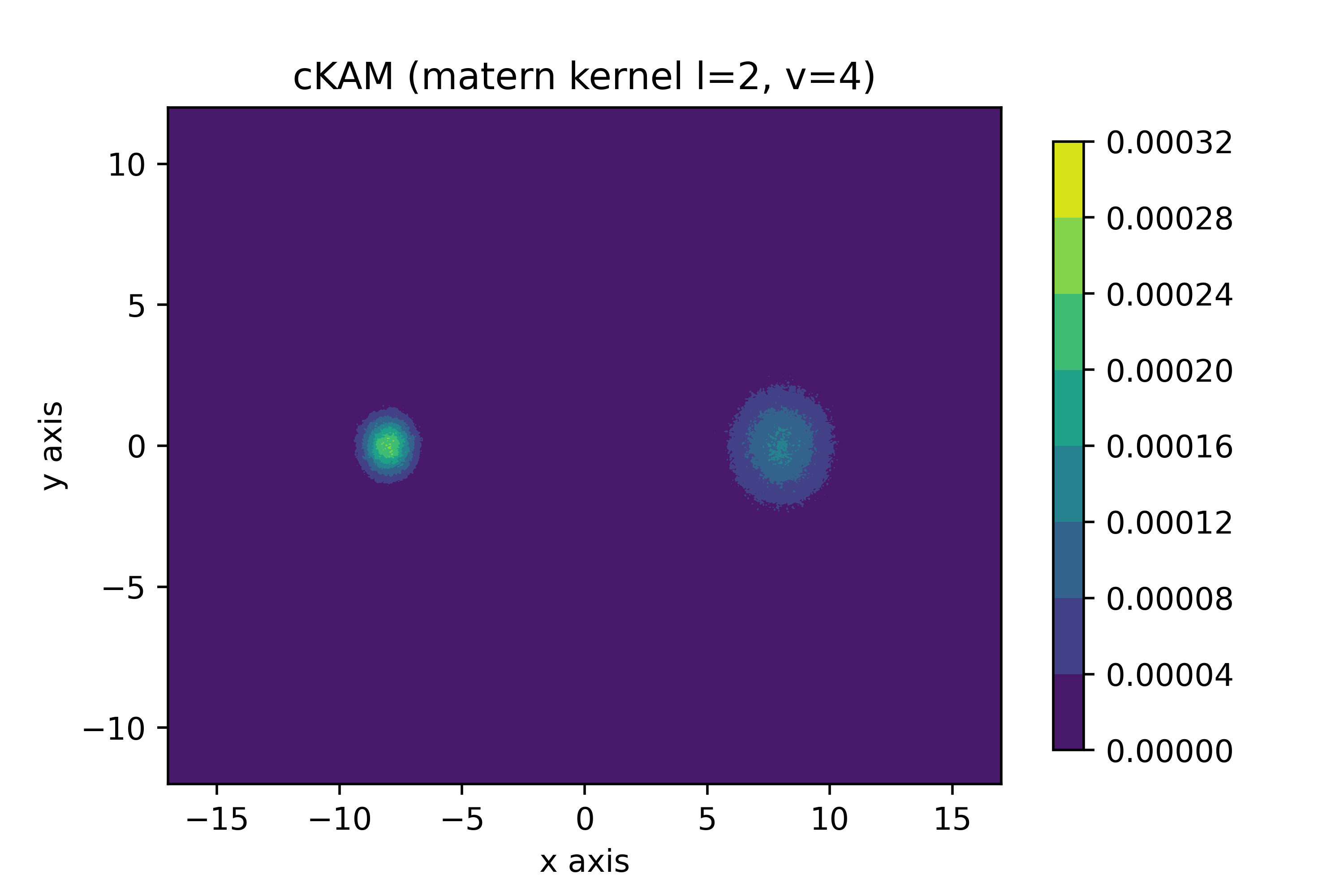}
\caption{\centering Density plot for all samplers}
\end{figure}

 We discovered empirically that any reasonable choice of kernel does not affect convergence rate for this experiment. We provide results of cKAM running with different kernel abd with all other hyperparameters/parameters kept the same: 
\vspace{-1.5em}
 \begin{figure}[H]
\centering
\includegraphics[width=0.55\textwidth]{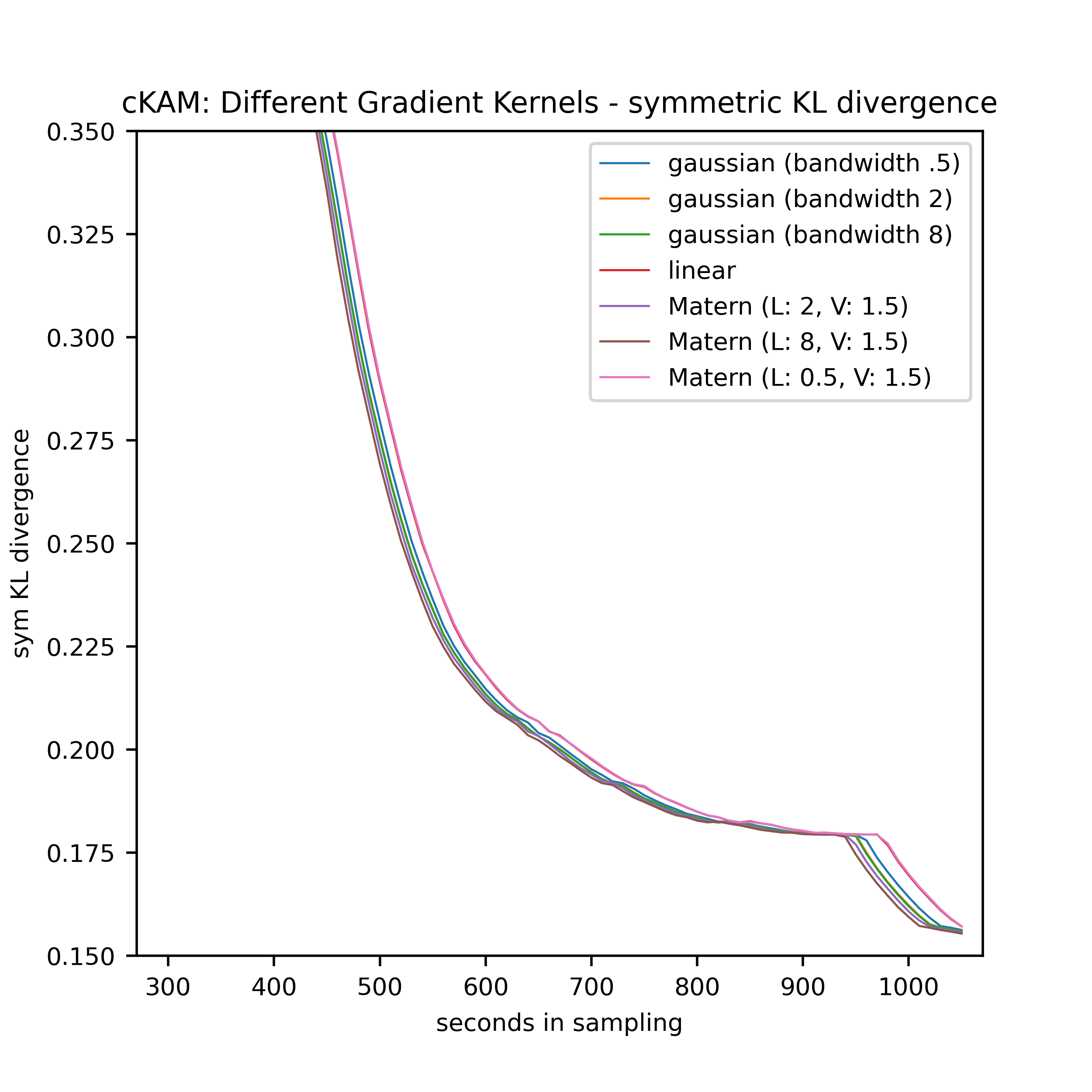} 
\caption{\centering KL divergence against time with different choices of kernel}
\end{figure}

 \subsection{2D Gaussian Mixture with increasing standard deviation}
 \label{std_appendix}
\begin{figure}[H]
\centering
\includegraphics[width=0.85\textwidth]{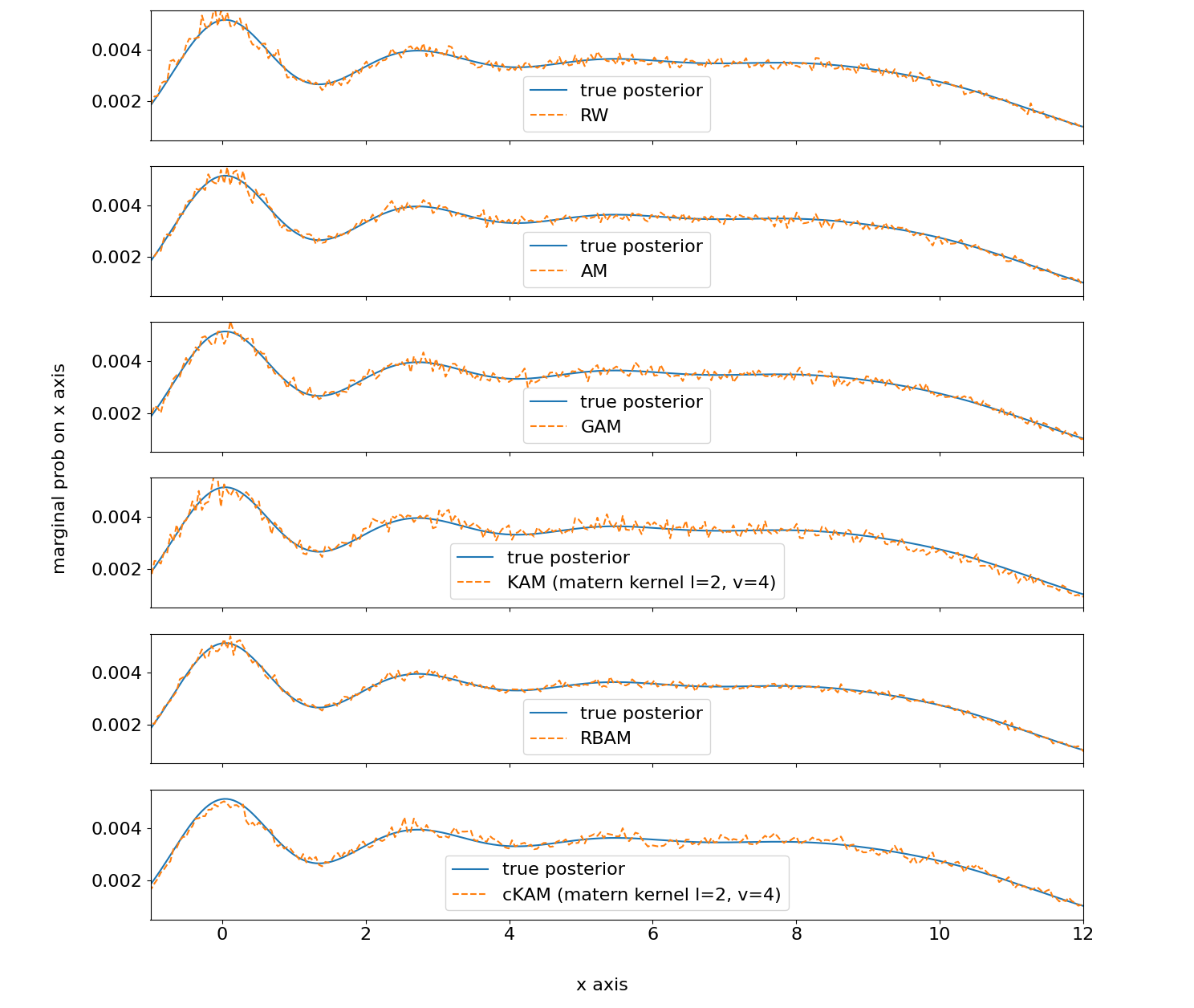}  
\label{fig:marginal_GM}
\caption{\centering Marginal probability on x-axis for all samplers}
\end{figure}

 We design the two-dimensional distribution composed of 5 Gaussian distributions with increasing standard deviation, leading to the true density plot in figure \ref{fig:std_posterior}. The analytical form of this distribution is:
\[
    \mathcal{P}(x) = \frac{1}{5} \sum_{i=1}^5 \mathcal{N}(x|\mu_i, \Sigma_i)
\]
 Where $\mu_i = 2.5\begin{bmatrix}i-1\\i-1\end{bmatrix}$ and $\Sigma_i = i^{1/2}\begin{bmatrix}1&0\\0&1\end{bmatrix}$. 
 We initialize all the samplers from $[0, 0]$, use a subsample size of 30 for KAM and 50 for cKAM, and run following hyperparameter settings for our samplers:\vspace{-4mm}
\begin{table}[H]
  \caption{Hyperparameter settings for 2D Gaussian Mixture distribution}
  \label{tab:hyp_bimodal}
  \centering
  \begin{tabular}{lcccccc}
    \toprule               
       & (initial) stepsize $\nu$ & RM stepsize $\eta$ & RM rate $\epsilon$   & $T/M$ &$\beta$ &optimal acceptance $\alpha^*$\\
    \midrule
    \vspace{.1cm}
    RW & $\sqrt{15}$ & N/A & N/A &N/A&N/A& N/A  \\
    \vspace{.1cm}
   AM & $\frac{2.38}{\sqrt{2}}$ & 0.01 & N/A &N/A& N/A  &N/A\\
    \vspace{.1cm}
    RBAM & N/A & 0.001 & N/A &N/A& N/A &N/A \\
    \vspace{.1cm}
    GAM &$\frac{2.38}{\sqrt{2}}$ & N/A & 0.75 & N/A &N/A  &0.234 \\
   KAM & $\frac{2*2.38}{\sqrt{2}}$ & N/A & 0.75 & N/A& N/A  &0.234 \\
    cKAM & $\frac{2*2.38}{\sqrt{2}}$ & N/A & 0.75 & 1000 & 0.4 &0.234 \\
    \bottomrule
  \end{tabular}
\end{table}
 Since the target distribution is symmetric by design, we also show the marginal probability on the dimension 1 in figure \ref{fig:marginal_GM}. 
 \newpage

\subsection{High-Dimensional Gaussian Mixture}
\label{high-GM}
We design the 32-dimensional distribution to have $5^{32}$ modes. The analytical form of the distribution is:
\[
    \mathcal{P}(x) = \sum \mathcal{N}\left(x|\mu, \Sigma \right)
\]

Where $\mu= \{-30, 15, 0, 15, 30\}^{32}, \Sigma$ is a $32\times32$ identity matrix multiplied by $15$. \par
We initialize all the samplers from $[0, 0]$, use a subsample size of 100 for KAM and 100 for cKAM, and run following hyperparameter settings for our samplers:\vspace{-4mm}
\begin{table}[H]
  \caption{Hyperparameter settings for 2D Gaussian Mixture distribution}
  \label{tab:hyp_bimodal}
  \centering
  \begin{tabular}{lcccccc}
    \toprule       
       & (initial) stepsize $\nu$ & RM stepsize $\eta$ & RM rate $\epsilon$   & $T/M$ &$\beta$ &optimal acceptance $\alpha^*$\\
    \midrule
    \vspace{.1cm}
    RW & $\frac{2.38}{\sqrt{32}}$ & N/A & N/A &N/A&N/A& N/A  \\
    \vspace{.1cm}
   AM & $\frac{2.38}{\sqrt{32}}$ & 0.01 & N/A &N/A& N/A  &N/A\\
    \vspace{.1cm}
    RBAM & N/A & 0.001 & N/A &N/A& N/A &N/A \\
    \vspace{.1cm}
    GAM &$\frac{2.38}{\sqrt{32}}$ & N/A & 0.75 & N/A &N/A  &0.234 \\
   KAM & $\frac{2*2.38}{\sqrt{32}}$ & N/A & 0.75 & N/A& N/A  &0.234 \\
    cKAM & $\frac{2*2.38}{\sqrt{32}}$ & N/A & 0.75 & 8000 & 0.6 &0.234 \\
    \bottomrule
  \end{tabular}
\end{table}


\end{document}